\documentclass{article}
\usepackage{spconf,amsmath,graphicx}
\usepackage{times}
\usepackage{epsfig}
\usepackage{amssymb}

\usepackage{tabularx}
\usepackage{pifont}
\usepackage{adjustbox}
\usepackage{multirow}
\usepackage{setspace}
\usepackage[caption=false,font=footnotesize]{subfig}
\usepackage{amsmath,scalerel}

\title{FAPM: Fast Adaptive Patch Memory for \\Real-time Industrial Anomaly Detection\vspace{-0.2cm}}
\name{Donghyeong Kim$^{1}$ \quad
   Chaewon Park$^{1}$ \quad
   Suhwan Cho$^{1}$ \quad
   Sangyoun Lee$^{1,2}$ \vspace{-0.2cm} \thanks{This work was supported by the KIST Institutional Program (ProjectNo.2E32283-23-064) and the Yonsei University Research Fund of 2021 (2021-22-0001).}}

\address{$^{1}$Yonsei University, Seoul, Korea
\\$^{2}$Korea Institute of Science and Technology (KIST)
\\{\tt\small \{2donghyung87, chaewon28, chosuhwan, syleee\}@yonsei.ac.kr} \vspace{-0.2cm}
}
\begin{document}
%

\maketitle

\setlength{\textfloatsep}{7pt}
	\begin{abstract}
	    Feature embedding-based methods have shown exceptional performance in detecting industrial anomalies by comparing features of target images with normal images. However, some methods do not meet the speed requirements of real-time inference, which is crucial for real-world applications. To address this issue, we propose a new method called Fast Adaptive Patch Memory (FAPM) for real-time industrial anomaly detection. FAPM utilizes patch-wise and layer-wise memory banks that store the embedding features of images at the patch and layer level, respectively, which eliminates unnecessary repetitive computations. We also propose patch-wise adaptive coreset sampling for faster and more accurate detection. FAPM performs well in both accuracy and speed compared to other state-of-the-art methods.
	    
	\end{abstract}
	
 \vspace{-0.4cm}
	\section{Introduction}
 \vspace{-0.3cm}
    Anomaly detection refers to the detection and localization of abnormal regions of images. Such a task is widely used in automated industries such as quality control, medical treatment, and surveillance. In particular, it is in high demand for estimating the abnormality of industrial products. Many recent studies~\cite{differnet,gan,itad,Padim,patchcore,patchsvdd,SPADE,TSDN} have investigated automated anomaly detection using computer vision.

    One factor that makes this task challenging is imbalanced data, which occurs when abnormal training data is more difficult to acquire than normal data because they rarely occur in the real world~\cite{survey,survey2}. Therefore, datasets are usually built using a one-class classification scheme, providing only normal data for training and both normal and abnormal data for testing. This is called an unsupervised approach for anomaly detection~\cite{survey2}. For such a scenario, the unsupervised network must learn the features of the unlabeled normal data during training and filter out the outlying features in the abnormal test data. Autoencoders~\cite{itad,AE,trustmae} and generative adversarial networks~\cite{gan, itad,mvtec,vevae, hou2021divide} have been explored for this purpose. However, when there is a high degree of intra-class variance, these methods encounter difficulty in producing clear reconstruction images. Moreover, it is hard to train a network for all the numerous categories of industrial objects. Therefore, many feature matching-based methods~\cite{Padim,patchcore,SPADE,STPM} have been proposed recently. Such methods~\cite{Padim,SPADE} predict abnormal regions by estimating the distance between feature distributions. Roth \textit{et al.}~\cite{patchcore} demonstrated outstanding performance by employing a memory bank to store the normal data embedding features extracted from a pretrained encoder.
    
    	\begin{figure}[!t]
		\begin{center}
			\includegraphics[width=0.9\columnwidth]{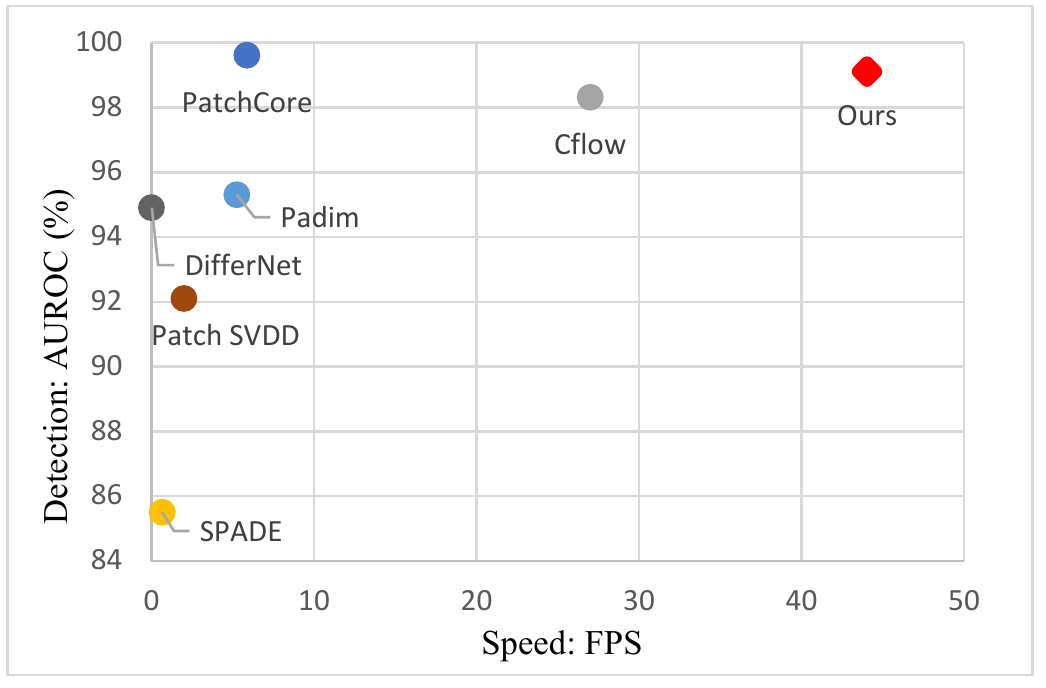}
		\end{center}
		\vspace{-0.6cm}
		\caption{Comparison of evaluation speed in frames per second (FPS) and image-level area under the receiver operating curve (AUROC) (\%) on the MVTecAD test set. The figure shows the values reported in the paper.}
  \vspace{-0.15cm}
		\label{speed}
	\end{figure}
	
    \begin{figure*}[!ht]
		\begin{center}
			\includegraphics[width=0.9\linewidth]{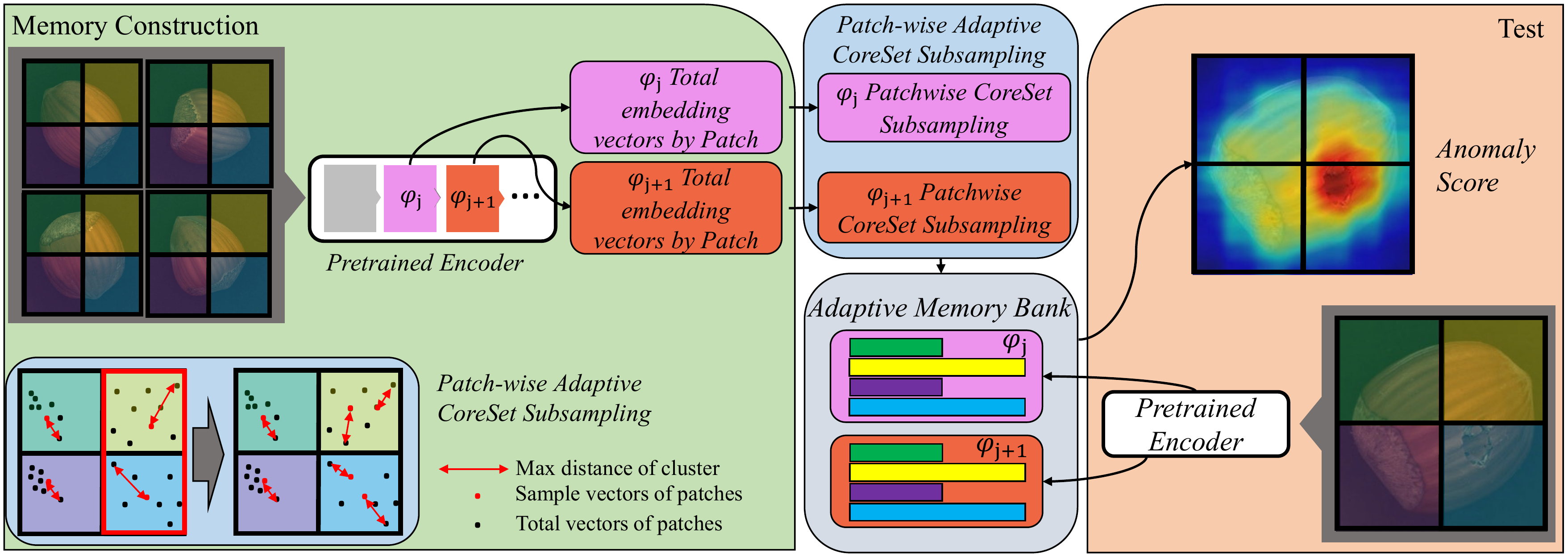}
		\end{center}
		\vspace{-0.8cm}
		\caption{An overview of Fast Adaptive Patch Memory. Each middle layer’s nominal feature vectors are saved into a patch in memory that represents the region of the image. To reduce the inference time and address the patch vector distribution imbalance, this patch memory bank has downsampled vectors at different ratios by each patch cluster’s distance score.}
        \vspace{-0.55cm}
		\label{architecture}
	\end{figure*}

    Inference speed is another critical issue for anomaly detection. For implementations in real-world industries, immediate anomaly detection is just as important as accurate detection. However, despite their outstanding accuracy, many feature matching-based methods~\cite{Padim,patchcore,SPADE} do not run sufficiently quickly or in real-time. Even though the objects are usually aligned in industrial situations, previous methods~\cite{patchcore,SPADE} are not run quickly enough or real-time because they calculate the similarity between all the embedding features of normal images with the target features, making the framework computationally expensive.

    To address these issues, we propose a method called Fast Adaptive Patch Memory (FAPM). FAPM uses a patch-wise memory bank, assuming that patches in the same location of each industrial image are likely to have similar embedding feature vectors. By comparing the features of patches in similar locations during inference, our method significantly reduces the computational cost. FAPM further increases the running speed by storing the features of each layer separately rather than upsampling and concatenating the features of deep layers. This avoids upsampling deeper feature vectors and reduces unnecessary calculations between the memory vectors and targets. For even better inference speed, we propose adaptive coreset sampling~\cite{core} to reduce the number of vectors stored in memory while retaining important vectors. Each patch in memory has a different embedding feature vector distribution, so flexible sampling is needed. We change the sampling ratio of each patch via the distribution of the embedding feature vectors. As shown in Fig.~\ref{speed}, our proposed FAPM was validated on a popular benchmark, performing well for real-time anomaly detection.
 
 	\vspace{-0.4cm}
	\section{Proposed Method}
 \vspace{-0.3cm}
 Fig.~\ref{architecture} provides an overview of FAPM. This section explains FAPM in three stages: 1) FAPM extracts features locally according to the patch and layer, 2) we select coreset embedding vectors by the appropriate sampling ratio based on the distribution of features for each patch and layer, and 3) FAPM detects the abnormal parts by comparing the features of the target with the memory bank’s downsampled vectors.

 	\vspace{-0.4cm}
	\subsection{Memory Construction}
    \vspace{-0.2cm}
	\label{section2.1}
	 \textbf{Patch-wise memory}~~Industrial anomaly datasets are primarily composed of roughly aligned images, which means similar information exists at the same location in each object class. Based on this assumption, the FAPM saves each patch’s embedding features separately at training time rather than indiscriminately saving all the image features in an integrated memory bank. Following, FAPM uses a pretrained encoder from ImageNet~\cite{imagenet} to extract the features of the nominal image $x$. We do not train the network because the pretrained encoder can generalize to most classes of objects.

The extracted features of $x$ can be split into $N_p$ grid-patches, where a patch $p_i~ (i\in\left\{1,2\dots,{N_p} \right\})$ is represented by a fixed number of $N_v$ feature vectors, where $\left\{v_{1}^{i},v_{2}^{i},\dots, v_{N_v}^{i} \right\}\in V_i$. These feature vectors are saved by patch-level memory $m_i$. It saves all $N$ images' specific location feature vectors $ \left\{V^{1}_i,V^2_{i}\dots V^N_i \right\}\in m_i$. $N$ is the number of total nominal images. All memory locations are gathered to form a patch-wise memory bank $\left\{m_1,m_2\dots,m_{N_p} \right\}\in M$, in which each $m_i$ represents a specific region of the $N$ training images.  The locally aware feature vectors $V_i$ of the nominal images are stored in the correct location for the patch memory $m_i$.

	\vspace{0.2cm}
\noindent \textbf{Layer-wise memory}~~ Many existing embedding features matching based methods include PatchCore~\cite{patchcore} and Padim~\cite{Padim} concatenate mid-level features, which avoids being overly generic or too heavily biased towards ImageNet~\cite{imagenet} classification ~\cite{patchcore}. Concatenating these features has the advantage of being able to aggregate the features from multiple layers at the same time. However, this naive concatenation leads to unnecessary upsampling of the deeper layer's features, which increases the vector size of the memory bank.

 To avoid such problem, FAPM uses separate layer-wise memory banks that store the feature vectors of each middle layer. We use the middle layers~$\mathbf{\left ( \psi_j,\psi_{j+1}, \dots   \right )}$ of pretrained model, except the layers at both ends. Each layer-wise memory bank is split into $N_p$ grids. $N_{v_{j}}$ is the value of $N_v$ that changes per layer. Our total memory banks from each layer are defined as $M_{\psi_j}\in \mathbb{R}^{N_p \times N \times N_{v_j}}.$

By avoiding concatenation, we decrease the size of the memory bank. This decrease helps to reduce the computational cost by reducing the number of unnecessarily upsampled elements of the embedding features.

 \vspace{-0.4cm}
	\subsection{Patch-wise Adaptive Coreset Sampling}
\begin{figure}[t]
	\centering
	\includegraphics[width=1\linewidth]{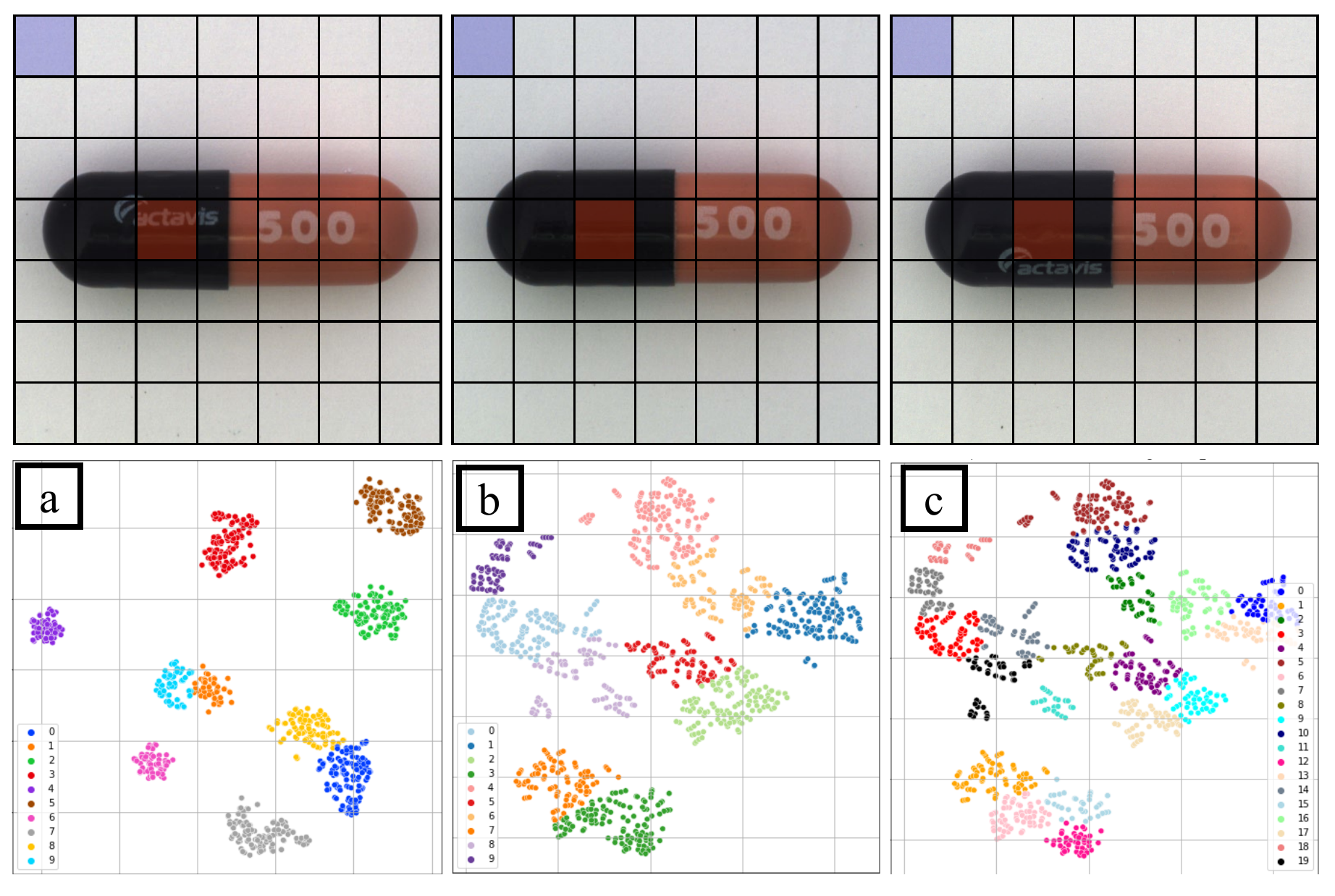}
	\vspace{-0.9cm}
	\caption{T-distributed stochastic neighbor embedding (T-SNE)~\cite{TSNE} of feature vectors by different patches of Capsule: (a) clusters ($K=10$) of blue patch memory vectors; (b) clusters ($K=10$) of red patch memory vectors; and (c) clusters ($K=20$) of red patch memory vectors.
}
	\vspace{-0.1cm}

	\label{patchvectors}
    \end{figure}

	\vspace{-0.2cm}
	For a fast inference speed, we select the most representative feature vectors from $m_i$. Following~\cite{coreset1,coreset2} and~\cite{patchcore}, we use a \textit{minimax facility location} greedy coreset selection~\cite{coreset1} to select the number of $K$ appropriate patch key vectors $V^{'}_{i}$ that are downsampled by each $V_i$. The key vectors $\left\{v^{'}_{i_1},v^{'}_{i_2},\dots,v^{'}_{i_K}\right\} \in V^{'}_{i}$ are the center vector of a cluster that groups similar vectors in $m_i$. As shown in Fig.~\ref{patchvectors}.(a) and (b), even with the same class of images, the distributions of feature vectors are different. Therefore, when simply using greedy coreset sampling~\cite{core}, each patch has a different distribution of key vectors.

  We propose patch-wise adaptive coreset sampling to avoid an imbalanced effect of the patches. First, we select the key vectors $V^{'}_{i}$ in $m_i$ using coreset sampling. This divides the vectors in $m_i$ by cluster with the center vector $v^{'}_{i_k}, ~k\in \left\{ 1,2,\dots, K \right\}$. Then, we estimate the size of each cluster by computing the Euclidean distance  $d_{i_k}$ between the outermost vector $v^{far}_{i_k}$ and $v^{'}_{i_k}$. Lastly, we find the largest score $d_{i_{max}}=max(d_{i_{k}})$ among all clusters in $m_i$.
  \vspace{-0.2cm}
 \begin{equation}
d_{i_k} =  1-\frac{2}{1+exp\left ( \left\| v_{i_k}^{far}-v_{i_k}^{'}\right\|^2 \right )}
\vspace{-0.2cm}
 \end{equation}
 Only for the patch memories $m_i$ whose $d_{i_{max}}$ is larger than a threshold $D_{th}$, coreset selection is performed once again by increasing the value of $K$. Then, we store the key vectors which selected by second coreset sampling. Each patch memory leaves a different number of key vectors $V^{'}_{i}$ according to its vector distribution. Patches with a large vector distribution leave more key vectors, and patches with a small distribution leave fewer key vectors. Fig.~\ref{patchvectors}. (b) and (c) show how this method affects the vector distribution imbalance problem. In this way, we are able to solve the vector distribution imbalance per patch to some extent.

\vspace{-0.4cm}
\subsection{Anomaly Detection}
\vspace{-0.15cm}
\noindent {\bf Anomaly detection with FAPM}~~Sampled memory $M^{'}_{\psi_{j}}$ saves the patch-wise key vectors $\left\{V^{'}_{j,1},V^{'}_{j,2},\dots,V^{'}_{j,{N_p}}\right\}\in V_{j}^{'}$ of the target class. At test time, target image $\hat{x}$ is input into the pretrained encoder that extracts the mid-level features $\hat{V_{j}}$. FAPM calculates the Euclidean distances between all patch-level vectors in the test image's patch $\hat{V}_{j,i}$ and memory's sampled vectors $V_{j,i}^{'}$ in $m_i$. After calculating all the Euclidean distances, we find the minimum. If the distance from the closest memory vector is too far, this test vector is considered abnormal. Some of the largest-distance vectors are used to calculate the image score. Then, we unify these maximum layer distances for estimating the image-level anomaly score. Like ~\cite{patchcore}, we leverage multiple feature vectors in memory to minimize the effects of outlier feature vectors.

Anomaly localization is computed, like ~\cite{Padim, patchcore}, by realigning computed vectors’ anomaly scores based on the spatial location. We also use Gaussian blurring with kernel size of $\sigma = 4$ to smooth the results.

\noindent {\bf Comparison with PatchCore}~~PatchCore~\cite{patchcore} saves and downsamples the number of $N_V$ concatenated mid-level features of whole nominal images to make a single memory that represents each object class. As all features are stored in a single memory, it does not include spatial information. In addition, PatchCore~\cite{patchcore} calculates the distances of all the test image’s number of $N_t$ mid-level features with the sampled vectors in memory. PatchCore's number of operations is $N_V \times N_t$. FAPM repeats the calculation by the number of patches $N_p$, but the number of vectors is reduced by the number of patches in both test images and memory vectors. Consequently, FAPM’s computational cost is $N_p\times \frac{N_V}{N_p} \times \frac{N_t}{N_p}.$

Moreover, PatchCore~\cite{patchcore} calculates the anomaly scores using two different feature levels. Higher-level features are first upsampled for resolution consistency with lower-level features. Then, the lower-level and higher-level features are concatenated and used for score calculation as the final features. However, this causes inefficient computation, as four times more calculations are required for the higher-level features due to the upsampling process. To prevent this, we decompose this redundant process into two steps: 1) separating the score calculation and 2) unifying the scores. Abnormality scores are first calculated for each level independently, and then the scores obtained are unified by weighted sum. The proposed layer-wise process operates identically to the single-step approach while requiring a significantly lower computational cost. Using these methods allows FAPM to have a lower computational cost than PatchCore~\cite{patchcore}.
\begin{table*}[]
\small
\centering
\resizebox{0.9\textwidth}{!}{%
\begin{tabular}{cc|c|c|c|c|c|c|c}
\multicolumn{2}{c|}{Class}                              & SPADE~\cite{SPADE}       & DifferNet~\cite{differnet} & PatchSVDD~\cite{patchsvdd}  & Padim~\cite{Padim}       & CFlow~\cite{cflow}        & PatchCore~\cite{patchcore}   & FAPM          \\ \hline
\hline
\multicolumn{1}{c|}{\multirow{5}{*}{texture}} & carpet     & (98.6, 97.5) & (84.0, -)  & (92.9, 92.6) & (-, 99.1)    & (\textbf{100}, \textbf{99.3})   & (98.7, 98.9) & (99.3, 98.9)   \\
\multicolumn{1}{c|}{}                         & grid       & (\textbf{99.0}, 93.7) & (97.1, -)  & (94.6, 96.2) & (-, 97.3)    & (97.3, \textbf{99.0})  & (98.2, 98.7) & (98.0, 97.8)   \\
\multicolumn{1}{c|}{}                         & leather    & (99.5, 97.6) & (99.4, -)  & (90.9, 97.4) & (-, 99.2)    & (97.7, \textbf{99.7})  & (\textbf{100}, 99.3)  & (\textbf{100}, 99.0)    \\
\multicolumn{1}{c|}{}                         & tile       & (89.8, 87.4) & (92.9, -)  & (97.8, 91.4) & (-, 94.1)    & (98.7, \textbf{98.0})  & (98.7, 95.6) & (\textbf{99.4}, 95.2)   \\
\multicolumn{1}{c|}{}                         & wood       & (95.8, 88.5) & (99.8, -)  & (96.5, 90.8) & (-, 94.9)    & (\textbf{99.6}, \textbf{96.7})  & (99.2, 95.0) & (99.3, 94.0)   \\ \hline
\multicolumn{1}{c|}{\multirow{10}{*}{object}} & bottle     & (98.1, 98.4) & (99.0, -)  & (98.6, 98.1) & (-, 98.3)    & (\textbf{100}, \textbf{99.0}) & (\textbf{100}, 98.6)  & (\textbf{100}, 98.2)    \\
\multicolumn{1}{c|}{}                         & cable      & (93.2, 97.2) & (86.9, -)  & (90.3, 96.8) & (-, 96.7)    & (\textbf{100}, 97.6) & (99.5, 98.4) & (99.5, \textbf{98.5})   \\
\multicolumn{1}{c|}{}                         & capsule    & (98.6, \textbf{99.0}) & (88.8, -)  & (76.7, 95.8) & (-, 98.5)    & (\textbf{99.3}, \textbf{99.0})  & (98.1, 98.8) & (98.6, \textbf{99.0})   \\
\multicolumn{1}{c|}{}                         & hazelnut   & (98.9, \textbf{99.1}) & (99.1, -)  & (92.0, 97.5) & (-, 98.2)    & (96.8, 98.9)  & (\textbf{100}, 98.7)  & (\textbf{100}, 98.6)    \\
\multicolumn{1}{c|}{}                         & metal nut  & (96.9, 98.1) & (95.1, -)  & (94.0, 98.0) & (-, 97.2)    & (91.9, \textbf{98.6})  & (\textbf{100}, 98.4)  & (\textbf{100}, 98.2)    \\
\multicolumn{1}{c|}{}                         & pill       & (96.5, 96.5) & (95.9, -)  & (86.1, 95.1) & (-, 95.7)    & (\textbf{99.9}, \textbf{99.0})  & (96.6, 97.1) & (96.0, 98.0)   \\
\multicolumn{1}{c|}{}                         & screw      & (99.5, 98.9) & (99.3, -)  & (81.3, 95.7) & (-, 98.5)    & (\textbf{99.7}, 98.9)  & (98.1, \textbf{99.4}) & (95.2, 99.0)   \\
\multicolumn{1}{c|}{}                         & toothbrush & (98.9, 97.9) & (96.1, -)  & (100, 98.1)  & (-, 98.8)    & (95.2, \textbf{99.0})  & (\textbf{100}, 98.7)  & (\textbf{100}, 98.7)    \\
\multicolumn{1}{c|}{}                         & transistor & (81.0, 94.1) & (96.3, -)  & (91.5, 97.0) & (-, 97.5)    & (99.1, 98.0)  & (\textbf{100}, 96.3)  & (\textbf{100}, \textbf{98.2})    \\
\multicolumn{1}{c|}{}                         & zipper     & (98.8, 96.5) & (98.6, -)  & (97.9, 95.1) & (-, 98.5)    & (98.5, \textbf{99.1})  & (98.8, 98.8) & (\textbf{99.5}, 98.6)   \\ \hline
\multicolumn{1}{l|}{}                         & Avg     & (85.5, 96.0) & (94.9, -)  & (92.1, 95.7) & (97.9, 97.5) & (98.3, \textbf{98.6})  & (\textbf{99.1}, 98.1) & (99.0, 98.0)   \\ \hline
\multicolumn{1}{l|}{}                         & FPS        & 1.5         & 2.0       & 2.1         & 4.4         & 27           & 5.9 ($23.4^*$)   & \textbf{44.1}
\end{tabular}%
}
\vspace{-0.3cm}
\caption{ Anomaly detection and segmentation performance on MVTec AD~\cite{mvtec} (image-level AUROC, pixel-level AUROC). We cite the values recorded in the paper. $*$ indicates the PatchCore implemented on our system for fair speed comparison.}
\vspace{-0.5cm}
\label{tab:result}
\end{table*}
\vspace{-0.4cm}
\begin{table}[]
\resizebox{\columnwidth}{!}{%
\begin{tabular}{l|ccc|c|c|c}
\hline
  & \multicolumn{1}{l}{Patch-wise} & \multicolumn{1}{l}{Layer-wise} & \multicolumn{1}{l|}{Adaptive sampling} & \multicolumn{1}{l|}{Img AUROC} & \multicolumn{1}{l|}{Pixel AUROC} & \multicolumn{1}{l}{FPS} \\ \hline
A &                                &                                &                                        & 98.9                           & \textbf{98.1}                    & 23.4                    \\
B & \checkmark                     &                                &                                        & 98.5                           & 97.5                             & 34.1                    \\
C &                    & \checkmark                                 &                                        & 98.3                           & 97.1                             & 35.7                    \\
D & \checkmark                     & \checkmark                     &                                        & 98.4                          & 97.8                             & \textbf{46.0}           \\
E & \checkmark                     & \checkmark                     & \checkmark                             & \textbf{99.0}                  & 98.0                             & 44.1                    \\ \hline
\end{tabular}%
}
\vspace{-0.4cm}
\caption{The impacts of our method. We show the image level AUROC ($\%$), pixel level AUROC ($\%$) and FPS.}
\vspace{-0.1cm}
\label{tab:abl}
\end{table}

	\vspace{-0.1cm}
	\section{Experiments}
	\vspace{-0.3cm}
	\subsection{Experimental Setup}
	\vspace{-0.15cm}
\noindent {\bf Implementation details}~~Experiments were performed on the MVTec Anomaly Detection benchmark~\cite{mvtec}. This dataset contains 3629 train images and 1725 test images in 15 industrial classes. There are ten object classes and five texture classes. We used center crop and normalization to the images. Our input image size was $224\times224$ pixels, we split this image into $N_p=49$ patches so our patch size was $32\times32$. We extracted mid-level features using Wide-ResNet50~\cite{WRN}’s second and third layers, which were pretrained by ImageNet~\cite{imagenet}. This network was also used by PatchCore~\cite{patchcore} and Padim~\cite{Padim}. We downsampled the memory vectors with a ratio of 10$\%$, and if $m_i$ had a higher $d_{i_{max}}$ than $D_{th}=0.5$, the ratio was doubled. At inference time, we considered four nearest memory vectors from each target vector. We do not use any normalization for real-time inference. We implemented our experiments with PyTorch Lightning using a single Nvidia Titan RTX 24GB. Our CPU was an Intel(R) Core(TM) i9-9900X CPU @ 3.50 GHz.

	\noindent {\bf Evaluation metric}~~For the evaluation, we adopted the area under the receiver operator curve (AUROC) obtained from the image-level scores. We also used AUROC for the pixel-level scores for anomaly localization. These metrics are used in most studies on image anomaly detection~\cite{differnet,Padim,patchcore,patchsvdd,SPADE,cflow}.
 \vspace{-0.4cm}
	\subsection{Experimental Results}
\begin{figure}[t]
	\centering
	\includegraphics[width=0.9\linewidth]{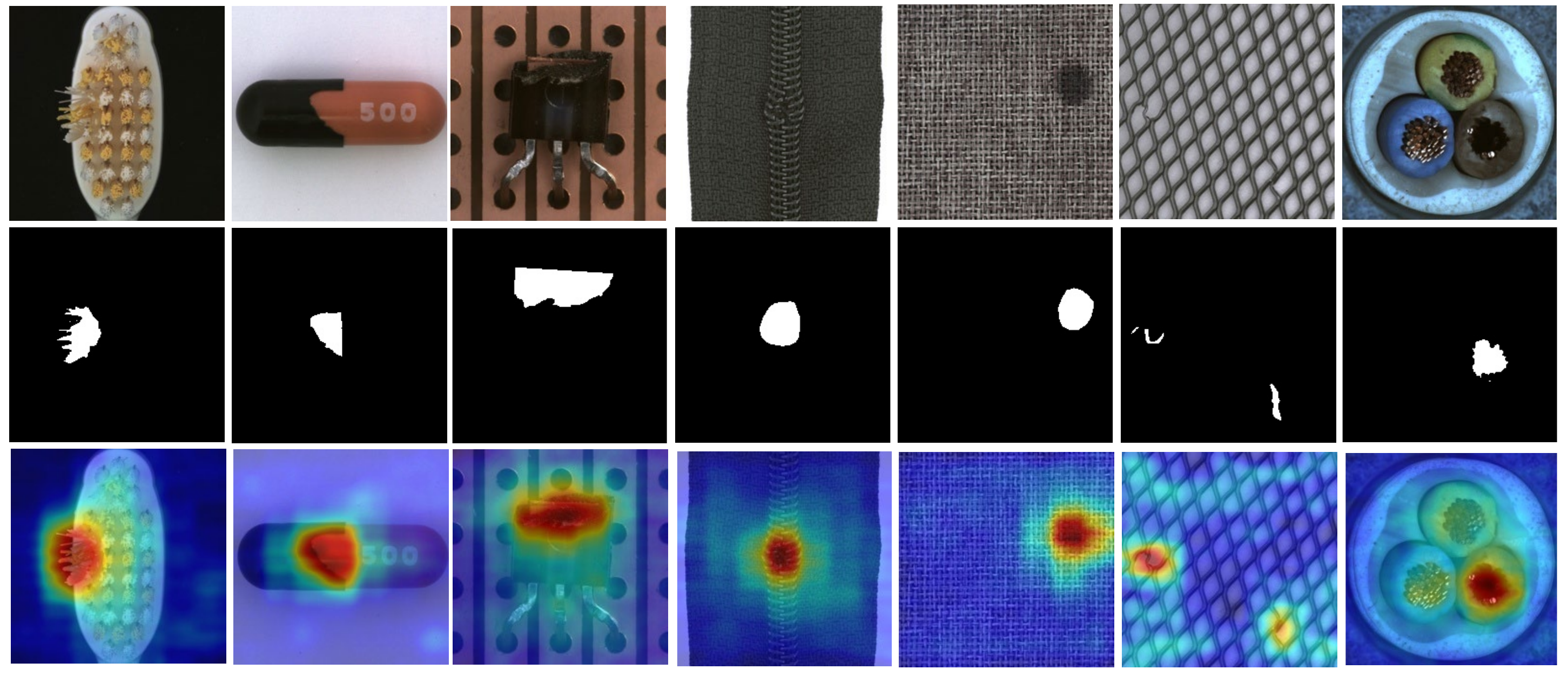}
	\vspace{-0.45cm}
	\caption{Anomaly Segmentation results of MVTec AD datasets. From top to bottom, input images, ground truth and anomaly segmentation heatmaps.}
    \vspace{-0.1cm}
	\label{anomaly_seg}
    \end{figure}

 \vspace{-0.2cm}
	Table~\ref{tab:result} shows the estimated performance of our FAPM on MVTec AD~\cite{mvtec}. The results show that FAPM was faster than state-of-the-art methods~\cite{differnet,Padim,patchcore,patchsvdd,SPADE,cflow}. FAPM was nearly twice as fast as PatchCore~\cite{patchcore} when implemented on our system. FAPM achieved 44.1 FPS demonstrating the ability for anomaly detection in real-time. Specifically, our method had the best detection and segmentation performance with some already-aligned categories such as transistor and zipper. Fig.~\ref{anomaly_seg} shows anomaly localization heatmaps where FAPM obtained more precise anomaly localization. Our method ran as fast as we intended and performed excellently.

 \vspace{-0.4cm}
	\subsection{Ablation Study}
	\label{ablation}
 \vspace{-0.2cm}
 Table~\ref{tab:abl} shows our method’s impact on anomaly detection and localization. We show the method’s performance in five experiments: (A) PatchCore-10\%~\cite{patchcore} on our system, where we only record its speed on our system; (B) using only patch-wise memory; (C) using only layer-wise memory; (D) using patch-wise and layer-wise memory with normal coreset sampling; and (E) our complete framework with adaptive coreset sampling. Table~\ref{tab:abl} (B), (C), and (D) show that our memory architecture divided into patch-wise and layer-wise memory improved the inference time. We implemented these experiments on GPU system, so inference time did not decrease linearly due to the computational cost decreasing, however all memory construction methods affected inference time in reality. Table 2 (E) shows that patch-wise adaptive sampling improved image anomaly detection by 0.6 percent points and segmentation performance by 0.2 percent points while maintaining the speed.

	\vspace{-0.5cm}
	\section{Conclusion}
	\vspace{-0.3cm}
	In this paper, we proposed a Fast Adaptive Patch Memory (FAPM) for industrial image anomaly detection. Our proposed method employs both patch-wise and layer-wise memory construction schemes for better utilization of memory architecture. We also proposed patch-wise adaptive coreset sampling by the distance score of each patch to solve the imbalance of patches. The proposed FAPM performed competitively on a benchmark dataset while showing exceptionally-fast inference speed in real-time.

\newpage
\newpage
\begingroup
\setstretch{0.85}
\bibliographystyle{IEEEbib}
\bibliography{refs}

\begin{thebibliography}{10}

\bibitem{differnet}
Marco Rudolph, Bastian Wandt, and Bodo Rosenhahn,
\newblock ``Same same but differnet: Semi-supervised defect detection with
  normalizing flows,''
\newblock in {\em Proceedings of the IEEE/CVF winter conference on applications
  of computer vision}, 2021, pp. 1907--1916.

\bibitem{gan}
Ta-Wei Tang, Wei-Han Kuo, Jauh-Hsiang Lan, Chien-Fang Ding, Hakiem Hsu, and
  Hong-Tsu Young,
\newblock ``Anomaly detection neural network with dual auto-encoders gan and
  its industrial inspection applications,''
\newblock {\em Sensors}, vol. 20, no. 12, pp. 3336, 2020.

\bibitem{itad}
Jonathan Pirnay and Keng Chai,
\newblock ``Inpainting transformer for anomaly detection,''
\newblock in {\em International Conference on Image Analysis and Processing}.
  Springer, 2022, pp. 394--406.

\bibitem{Padim}
Thomas Defard, Aleksandr Setkov, Angelique Loesch, and Romaric Audigier,
\newblock ``Padim: a patch distribution modeling framework for anomaly
  detection and localization,''
\newblock in {\em International Conference on Pattern Recognition}. Springer,
  2021, pp. 475--489.

\bibitem{patchcore}
Karsten Roth, Latha Pemula, Joaquin Zepeda, Bernhard Sch{\"o}lkopf, Thomas
  Brox, and Peter Gehler,
\newblock ``Towards total recall in industrial anomaly detection,''
\newblock in {\em Proceedings of the IEEE/CVF Conference on Computer Vision and
  Pattern Recognition}, 2022, pp. 14318--14328.

\bibitem{patchsvdd}
Jihun Yi and Sungroh Yoon,
\newblock ``Patch svdd: Patch-level svdd for anomaly detection and
  segmentation,''
\newblock in {\em Proceedings of the Asian Conference on Computer Vision},
  2020.

\bibitem{SPADE}
Niv Cohen and Yedid Hoshen,
\newblock ``Sub-image anomaly detection with deep pyramid correspondences,''
\newblock {\em arXiv preprint arXiv:2005.02357}, 2020.

\bibitem{survey}
Varun Chandola, Arindam Banerjee, and Vipin Kumar,
\newblock ``Anomaly detection: A survey,''
\newblock {\em ACM computing surveys (CSUR)}, vol. 41, no. 3, pp. 1--58, 2009.

\bibitem{survey2}
Xian Tao, Xinyi Gong, Xin Zhang, Shaohua Yan, and Chandranath Adak,
\newblock ``Deep learning for unsupervised anomaly localization in industrial
  images: A survey,''
\newblock {\em IEEE Transactions on Instrumentation and Measurement}, 2022.

\bibitem{AE}
Mayu Sakurada and Takehisa Yairi,
\newblock ``Anomaly detection using autoencoders with nonlinear dimensionality
  reduction,''
\newblock in {\em Proceedings of the MLSDA 2014 2nd workshop on machine
  learning for sensory data analysis}, 2014, pp. 4--11.

\bibitem{trustmae}
Daniel~Stanley Tan, Yi-Chun Chen, Trista Pei-Chun Chen, and Wei-Chao Chen,
\newblock ``Trustmae: A noise-resilient defect classification framework using
  memory-augmented auto-encoders with trust regions,''
\newblock in {\em Proceedings of the IEEE/CVF winter conference on applications
  of computer vision}, 2021, pp. 276--285.

\bibitem{mvtec}
Paul Bergmann, Michael Fauser, David Sattlegger, and Carsten Steger,
\newblock ``Mvtec ad--a comprehensive real-world dataset for unsupervised
  anomaly detection,''
\newblock in {\em Proceedings of the IEEE/CVF conference on computer vision and
  pattern recognition}, 2019, pp. 9592--9600.

\bibitem{vevae}
Wenqian Liu, Runze Li, Meng Zheng, Srikrishna Karanam, Ziyan Wu, Bir Bhanu,
  Richard~J Radke, and Octavia Camps,
\newblock ``Towards visually explaining variational autoencoders,''
\newblock in {\em Proceedings of the IEEE/CVF Conference on Computer Vision and
  Pattern Recognition}, 2020, pp. 8642--8651.

\bibitem{hou2021divide}
Jinlei Hou, Yingying Zhang, Qiaoyong Zhong, Di~Xie, Shiliang Pu, and Hong Zhou,
\newblock ``Divide-and-assemble: Learning block-wise memory for unsupervised
  anomaly detection,''
\newblock in {\em Proceedings of the IEEE/CVF International Conference on
  Computer Vision}, 2021, pp. 8791--8800.

\bibitem{STPM}
Guodong Wang, Shumin Han, Errui Ding, and Di~Huang,
\newblock ``Student-teacher feature pyramid matching for unsupervised anomaly
  detection,''
\newblock {\em arXiv preprint arXiv:2103.04257}, 2021.

\bibitem{core}
Pankaj~K Agarwal, Sariel Har-Peled, Kasturi~R Varadarajan, et~al.,
\newblock ``Geometric approximation via coresets,''
\newblock {\em Combinatorial and computational geometry}, vol. 52, no. 1, 2005.

\bibitem{imagenet}
Alex Krizhevsky, Ilya Sutskever, and Geoffrey~E Hinton,
\newblock ``Imagenet classification with deep convolutional neural networks,''
\newblock {\em Communications of the ACM}, vol. 60, no. 6, pp. 84--90, 2017.

\bibitem{coreset1}
Ozan Sener and Silvio Savarese,
\newblock ``Active learning for convolutional neural networks: A core-set
  approach,''
\newblock {\em arXiv preprint arXiv:1708.00489}, 2017.

\bibitem{coreset2}
Samarth Sinha, Han Zhang, Anirudh Goyal, Yoshua Bengio, Hugo Larochelle, and
  Augustus Odena,
\newblock ``Small-gan: Speeding up gan training using core-sets,''
\newblock in {\em International Conference on Machine Learning}. PMLR, 2020,
  pp. 9005--9015.

\bibitem{TSNE}
Laurens Van~der Maaten and Geoffrey Hinton,
\newblock ``Visualizing data using t-sne.,''
\newblock {\em Journal of machine learning research}, vol. 9, no. 11, 2008.

\bibitem{cflow}
Denis Gudovskiy, Shun Ishizaka, and Kazuki Kozuka,
\newblock ``Cflow-ad: Real-time unsupervised anomaly detection with
  localization via conditional normalizing flows,''
\newblock in {\em Proceedings of the IEEE/CVF Winter Conference on Applications
  of Computer Vision}, 2022, pp. 98--107.

\bibitem{WRN}
Sergey Zagoruyko and Nikos Komodakis,
\newblock ``Wide residual networks,''
\newblock in {\em Proceedings of the British Machine Vision Conference (BMVC)},
  Edwin R.~Hancock Richard C.~Wilson and William A.~P. Smith, Eds. September
  2016, pp. 87.1--87.12, BMVA Press.

\end{thebibliography}
\endgroup

\end{document}